\documentclass{article}
\usepackage{ijcai19}

\usepackage{times}
\usepackage{soul}
\usepackage{url}
\usepackage[hidelinks]{hyperref}
\usepackage[utf8]{inputenc}
\usepackage[small]{caption}
\usepackage{graphicx}
\usepackage{amsmath}
\usepackage{booktabs}
\usepackage{hyperref}
\usepackage{amsmath}
\usepackage{graphicx}
\usepackage{makecell}
\usepackage{booktabs,chemformula}
\usepackage{algorithm}
\usepackage[noend]{algpseudocode}

\usepackage[colorinlistoftodos]{todonotes}

\usepackage{amssymb}

\title{Automated Machine Learning with Monte-Carlo Tree Search}

\author{
Herilalaina Rakotoarison \and Marc Schoenauer \and Michèle Sebag\\
\affiliations
TAU, LRI-CNRS–INRIA, 
Université Paris-Saclay, France\\
\emails
\{herilalaina.rakotoarison, marc.schoenauer\}@inria.fr,
sebag@lri.fr
}

\begin{document}

\maketitle

\newcommand{\argmin}{\operatornamewithlimits{argmin}}
\newcommand{\argmax}{\operatornamewithlimits{argmax}}
\def\mosaic{{{\sc Mosaic}}}
\def\autosklearn{{{\sc Auto-Sklearn}}}
\def\as{{{\sc Auto-Sklearn}}}
\def\Hyperband{{{\sc Hyperband}}}
\def\AutoWeka{{{\sc Auto-Weka}}}
\def\Spearmint{{{\sc Spearmint}}}
\def\TPE{{{\sc Tpe}}}
\def\Tpot{{{\sc Tpot}}}
\def\TPOT{{{\sc Tpot}}}
\def\Alpha{{{\sc AlphaD3M}}}
\def\SMAC{{{\sc Smac}}}
\def\Zero{{{\sc AlphaGo Zero}}}
\def\Autostacker{{{\sc AutoStacker}}}
\def\C{{\mbox{$C_{ucb}$}}} 

\def\hatQ{Q_{\widehat F}}
\def\hatI{I_{\widehat F}}%
\def\EE{{\rm I\hspace{-0.50ex}E}}
\def\RR{{\rm I\hspace{-0.50ex}R}}
\def\hyper{hyper-parameters}
\def\hypernots{hyper-parameter}
\def\V{{\sc .v}}
\def\ML{{\sc .Ml}}

\def\a{\mbox{${\bf a}$}}
\def\s{\mbox{${\bf s}$}}
\def\x{\mbox{${\bf x}$}}
\def\z{\mbox{${\bf z}$}}
\def\kips{{$k$-ps}}
\def\kpips{{$k+1$-ps}}

\begin{abstract}
The AutoML task consists of selecting the proper algorithm in a machine learning portfolio, and its hyperparameter values, in order to deliver the best performance on the dataset at hand. \mosaic, a Monte-Carlo tree search (MCTS) based  approach, is presented to handle the AutoML hybrid structural and parametric expensive black-box optimization problem. Extensive empirical studies are conducted to independently assess and compare: i) the optimization processes based on Bayesian optimization or MCTS; ii) its warm-start initialization; iii) the ensembling of the solutions gathered along the search. \mosaic\ is assessed on the OpenML 100 benchmark and the Scikit-learn portfolio, with statistically significant gains over \as, winner of former international AutoML challenges.
\end{abstract}



\section{Introduction}
The automated selection of the machine learning (ML) algorithm yielding the best performance on the problem at hand, referred to as AutoML, has attracted interest since the late 1980s  \cite{Brazdil2018}: there exists no killer ML algorithm dominating all others on all datasets \cite{NFL}, and  ML algorithms demonstrate a high sensitivity w.r.t. their \hyper.  With the explosion of machine learning applications, the AutoML issue becomes even more acute. AutoML gradually extended to \hyper\ optimization 
\cite{bergstra2011algorithms}, and finally tackles the optimization of the overall ML pipeline from data preparation to model learning
\cite{NIPS2015_5872,li2016hyperband,OlsonGECCO2016,AutoStaker}. 
Several AutoML international challenges have been organized in the last decade \cite{guyon2015design,AutoML2018}, spurring the development of efficient AutoML systems such as \AutoWeka\ \cite{kotthoff_auto-weka_2017}, \Hyperband\ \cite{li2016hyperband}, \TPOT\ \cite{OlsonGECCO2016} and the challenge winner \as\ \cite{NIPS2015_5872} (more in section \ref{sec:soa}). 

AutoML systems tackle a black-box expensive optimization problem: For a given target dataset,
\begin{equation}
\mbox{Find~} \x^{*} \in \arg\max_{x \in X} {\cal F}(\x),
\label{eq:1}
\end{equation}
where $X$ is the structural and parametric space of ML configurations (containing categorical and continuous parameters with hierarchical dependencies), and ${\cal F}(\x)$ the performance of the model learned from the dataset at hand using configuration $\x$. ML configurations and pipelines are used interchangeably in the following.

A main difficulty of the AutoML optimization problem  lies in the search space: an ML pipeline is a series of components (algorithms), together with their own \hyper. The task thus consists in solving the combinatorial optimization of the pipeline structure, the performance of which depends on the parametric optimization of its component \hyper. 

Most AutoML approaches tackle both problems using a single optimization approach technique (e.g., Bayesian Optimization or Evolutionary Algorithms) whereas both problems are of very different nature. The contribution of the paper, presenting the \mosaic\ ({\em MOnte-Carlo tree Search for AlgorIthm Configuration}) approach,\footnote{\mosaic\ is publicly available under an open source license at \href{https://github.com/herilalaina/mosaic_ml}{https://github.com/herilalaina/mosaic\_ml}.} is to use the best approach for each problem while tightly coupling both optimizations (section \ref{sec:mosaic}).

Specifically, the optimization of the pipeline can be viewed as a sequential decision process; Monte-Carlo Tree Search (MCTS) \cite{kocsisMCTS} has demonstrated its ability to efficiently solve such sequential problems. On the other hand, Bayesian optimization \cite{EI,WangPhD} has been very successful solving expensive optimization problems, in particular in the context of \hyper\ tuning \cite{hutter2011sequential}. These two approaches are coupled in \mosaic\, and their coupling relies on a surrogate model of the performance of the pipelines, as in \as. However, this surrogate model is not only used to guide the local search of the \hyper, it is also incorporated at the heart of the MCTS search of the best pipeline structure.

The paper is organized as follows. Section \ref{sec:soa} 
discusses the state of the art in AutoML, and 
presents the MCTS formal background. Section \ref{sec:mosaic} gives a detailed overview of the proposed \mosaic\ approach. The experimental setting and the goals of experiments are presented in Section \ref{sec:goal}. Section  \ref{sec:expe} reports on the empirical validation\footnote{We warmly thank \as\ authors, who kindly provided many explanations together with their open source code. We also thank \TPOT\ authors, who provide an open source easy-to-use software package.}  of \mosaic\ on the OpenML benchmark suite and the Scikit-learn portfolio, demonstrating statistically significant gains over \as\ and \Tpot\ \cite{OlsonGECCO2016}.

\section{Related work}
\label{sec:etat-de-l-art}\label{sec:soa}
This section briefly reviews previous work on the {\em per-instance} AutoML problem (Eq. (\ref{eq:1})), first focusing on approaches using surrogate models and Bayesian Optimisation, then on MCTS and other approaches. Approaches focused on specific issues, e.g., neural architecture optimization \cite{wistuba_deep_2019}, are omitted due to space limitations.

\subsection{Surrogate Model-based optimization} \label{sec:sur}

Most prominent approaches today 
proceed iteratively, learning and exploiting an estimate of the optimization objective $\cal F$, called {\em surrogate model}.

\paragraph{Learning a surrogate model.}
At step $t$, surrogate model $\widehat{\cal F}_t: X \mapsto \RR$ is learned from the set $\{ (x_u,{\cal F}(x_u)), u=1 \ldots t\}$ gathering the previously selected configurations and their associated performances.  $\widehat{\cal F}_t$ is then used to determine the most promising candidate $x_{t+1}$, see below. 

As said, a main difficulty lies in the structure of space $X$. In all generality, this space includes categorical features (e.g., the name of the ML algorithm, the type of pre-processing) and continuous or integer features, the number and range of which depend on the value of the categorical features (e.g. the algorithm or pre-processing method). 
Diverse surrogate model hypothesis spaces have been considered:
the {\em Sequential Model-based Algorithm Configuration} (\SMAC) \cite{hutter2011sequential}, like \AutoWeka\  \cite{kotthoff_auto-weka_2017} and \autosklearn\ \cite{NIPS2015_5872}, are  based on Random Forests; \cite{bergstra2011algorithms} use a Tree-structure Parzen Estimator (\TPE); \Spearmint\ \cite{snoek2015scalable} is based on Gaussian processes (GP). An extensive comparison of these approaches \cite{eggensperger2013towards} shows that 
\SMAC\ and \TPE\ perform best for high dimensional and mixed hyperparameter optimization problems, while the GP-based \Spearmint\ performs best on low dimensional continuous search spaces.

\paragraph{Surrogate model-based optimization.} Surrogate models are often exploited along {\em Bayesian optimization} (BO) \cite{EI,WangPhD}. Assuming that model $\widehat{\cal F}_t$ yields the performance distribution for any given $\x$, the most promising $\x^*_{t+1}$ is determined by maximizing the expected improvement on the current best value ${\cal F}({\x_t^*})$ \cite{EI}, or more generally an acquisition function balancing performance expectation and variance \cite{WangPhD}. 

A simple alternative is to learn a surrogate model as a random forest, yielding both a performance estimate and a variance estimate for any configuration.
The next candidate $\x_{t+1}$ is the configuration maximizing the approximate acquisition function, out of a number of configuration samples. The key issue here is the distribution used to sample the configuration space. For instance,
\autosklearn, as it uses \SMAC, considers a small number of configurations close to the best-so-far configuration, augmented with a large number of uniformly sampled configurations.

\subsection{Monte-Carlo Tree Search} \label{mctsAlgorithm}
\begin{figure}
\includegraphics[scale=0.123]{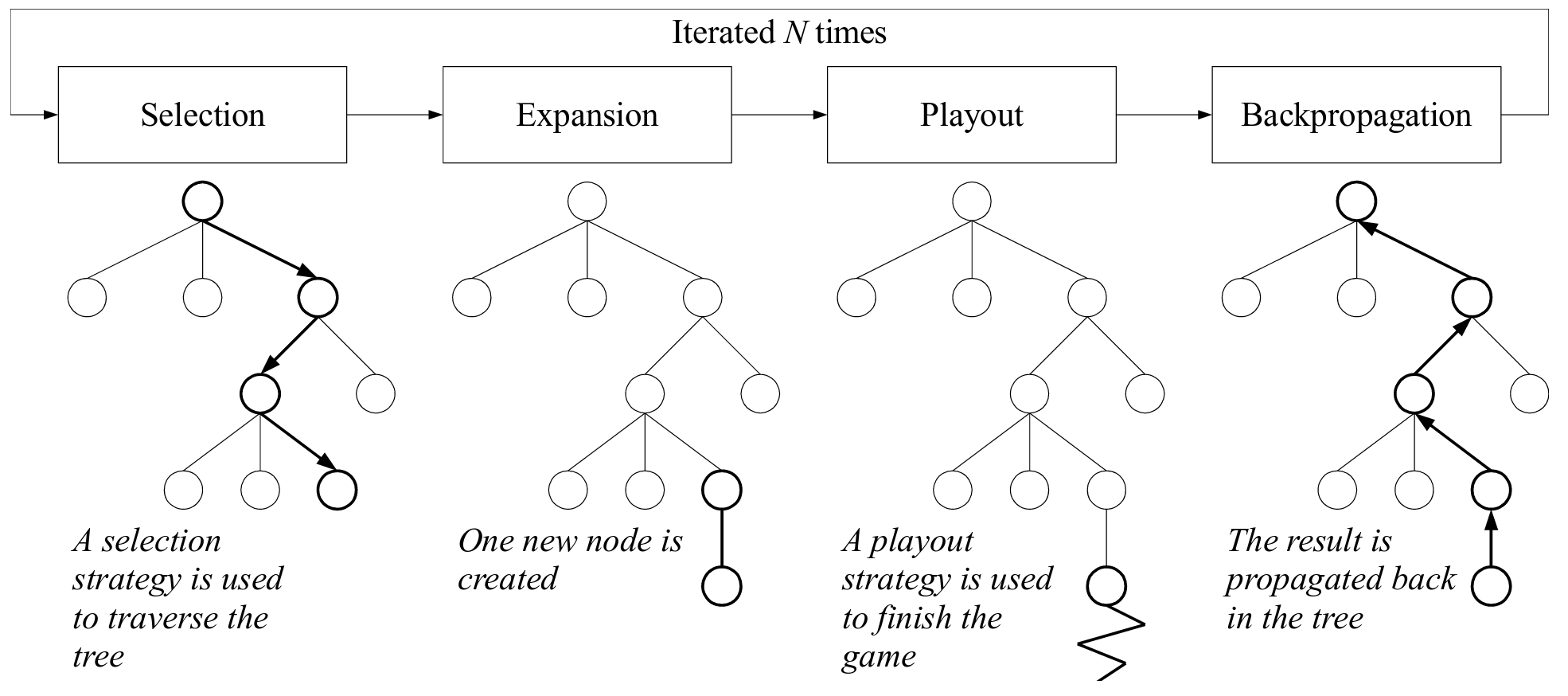}
\caption{Monte-Carlo Tree Search: each iteration involves four phases \protect\cite{Chaslot08}.}
\label{fig:MCTS_execution}
\end{figure}

An alternative to Bayesian optimization is based on Monte-Carlo Tree Search \cite{kocsisMCTS}. 
Considering a tree-structured search space $X$,  MCTS iteratively explores the space, gradually biasing the exploration toward the most promising regions of the search tree. 
Each iteration, referred to as tree-walk (Fig. \ref{fig:MCTS_execution}), involves four phases \cite{gellysilver}: 
\paragraph{Down the MCTS tree:} The first phase traverses the MCTS tree from the root node. In each (non-leaf) node $\s$ of the tree, the next node $\s.a$ to visit is classically selected among the child nodes of $\s$ using the multi-armed bandit Upper Confidence Bound criterion \cite{Auer02}:
\begin{equation}
\mbox{select } \arg\max_{a} \left\{ \hat{\mu}_{s.a} + \C \sqrt{\frac{\log{n(s)}}{n(s.a)}} \right\}
 \label{eq:ucb}
\end{equation}
with $\hat{\mu}_{s.a}$ the average reward gathered over all tree-walks with prefix $\s.a$, $n(\s)$ (resp. $n(\s.a)$) the number of visits to node $\s$ (resp. node $\s.a$), and $\C$ a problem-dependent constant that controls the exploitation {\em vs} exploration trade-off; 

\paragraph{Expansion:} When arriving at a leaf node, a new child node might be added.  The choice of the new node can be guided using e.g. the {\em Rapid Action Value Estimate} \cite{gellysilver}. The number of child nodes is controlled and gradually extended  along the Progressive Widening strategy \cite{couetoux13}: A new child node is added whenever the integer value of $n(s)^{PW}$ increases by one, $PW$ being a user-defined parameter (typically 0.6).

\paragraph{Playout:} After the expansion phase, a playout strategy is used to complete the tree-walk until reaching a terminal node and computing the associated reward; 
\paragraph{Back-propagation:} The reward value is back-propagated along the current path, incrementing $n(\s)$ for all visited nodes and updating $\hat{\mu}_s$ accordingly. 

Taking inspiration from \Zero\ \cite{AlphaGOZero}, the \Alpha\ system builds upon MCTS to explore the pipeline search space \cite{AlphaD3M}.  The difference compared to mainstream AutoML systems is twofold. Firstly, \Alpha\ explores the sequences of actions (insertion, deletion, replacement of pipeline parts) on pipelines, as opposed to directly exploring space $X$. Secondly, \Alpha\ learns (resp. exploits) a recurrent neural net to encode the action probability of success (resp. probability of selection) conditioned on the current state, {\em in lieu} of surrogate model or selection rule. 


\subsection{Expensive optimization}
As said, AutoML is an expensive black-box optimization problem:  computing ${\cal F}(\x)$ amounts to run the whole ML pipeline $\x$ on the considered dataset. 
Several approaches have been proposed to reduce the computational cost. A first one consists of sub-sampling the training dataset  \cite{swersky2014freeze,li2016hyperband,klein2016fast}. Two surrogate models are built in \cite{klein2016fast}: one for the performance reached depending on the configuration $\x$ and a fraction $\rho$ of the training set considered, another one for the actual computational cost of running $\x$ on a fraction $\rho$ of the data. Both models are jointly exploited to determine the most promising pipeline in terms of performance improvement and moderate computational cost. 

Another approach is \Hyperband\ \cite{li2016hyperband}, launching a large number of random candidate configurations, subject to a given cut-off time. \Hyperband\  iteratively prunes the unpromising candidates, and re-examines the other candidates with a larger cut-off, until the best candidates are allowed to run with no computational cost constraint. After its authors, \Hyperband\  outperforms \SMAC\ and \TPE\ for \hyper\ optimization on neural networks and support vector machines, though its performances are sensitive to its own hyper-\hyper.

Two evolutionary approaches (EAs) have been proposed, handling particular ML pipelines. \TPOT\ uses Genetic Programming to evolve pipelines made of parallel preprocessing and feature construction branches, that feed some model building method. A comparative study \cite{balaji_benchmarking_2018} reports that \TPOT\ is outperformed by \as\ on classification problems while the reverse is true on regression problems. 
\Autostacker\ \cite{AutoStaker} 
builds an ML pipeline by evolving new artificial features, and adding them to the original dataset. The whole stack is optimized using a vanilla EA with {\em ad hoc} mutation and crossover. \Autostacker\ outperforms \TPOT, and yields some better results than \as, though both algorithms have very different ways of handling CPU time.

\subsection{Search initialization and solution agregation}\label{metaAndEnsemble}
It is long known that initialization is a most critical step for ill-posed optimization problems. The selection of the first candidates $\x_u$ will govern the quality of the surrogate model (section \ref{sec:sur}) and the time-to-good configurations: the better the initial $\x_u$s, the more accurate the surrogate model will be {\em in the worthy part of the search space}\footnote{Moderate mistakes in the low-performing regions do not harm since these regions will not be much visited.}. 

The selection of the initial $\x_u$s in \as\ is based on the so-called {\em MetaLearning} heuristics. Formally, \as\ is provided with an archive, gathering pairs $(\z_i,\x_i)$ where the meta-feature\footnote{Meta-features are used to describe datasets, using statistical, information theoretic and landmark-based measures \cite{KateSmithMiles}.} vector $\z_i$ describes the $i$-th dataset and $\x_i$ is the best known pipeline for this dataset. 
Letting $\z$ denote the meta-feature vector associated to the current dataset, its nearest neighbors in the archive (in the sense of the Euclidean distance on the meta-feature vector space) are computed and
the $\x_i$s associated with these neighbors are used by \as\ as first configurations \cite{NIPS2015_5872}.   

Finally, the sequence of solutions found by an AutoML process can be exploited in the spirit of ensemble learning \cite{caruana2004ensemble}. \as.Ensemble delivers the compound model defined as the weighted sum of the models learned along the search, where the weights are optimized on a validation set. 





\section{MCTS-aided Algorithm Configuration}
\label{sec:mosaic}
After introducing some notations, this section presents \mosaic\ and discusses its components.

An ML pipeline $\x$ involves a fixed ordered sequence of $\ell$ decisions, respectively selecting the data preprocessing (including categorical variable encoding, missing value imputation, rescaling), feature  selection, and learning algorithms. At the $i^{th}$ decision step, some algorithm $a_i \in {\cal A}_i$ is selected (with ${\cal A}_i$ the finite set of possible algorithms at $i^{th}$ step). Denoting $\Theta (a_i)$ the (possibly varying dimension) space of \hyper\ associated with $a_i$, the eventual pipeline is described as $\x = (a_1,\theta_1), \ldots (a_\ell, \theta_\ell)$, with $\theta_i \in \Theta(a_i)$. A complete {\em pipeline structure} is an\footnote{Note that elements in ${\cal A}$ are not all admissible. Domain knowledge is used to early discard the non-admissible sequences $a_1 \ldots a_i$.} $\ell$-uple $\a = (a_1, \ldots a_\ell) \in {\cal A} = {\cal A}_1 \times \ldots \times {\cal A}_\ell$, with $\Theta(\a) = \Theta(a_1) \times \ldots \times \Theta(a_\ell)$ its associated \hypernots\ space. A $k$-pipeline structure (\kips) is a $k$-tuple $\s = (a_1, \ldots a_k) \in {\cal A}_1 \times \ldots \times {\cal A}_k$, with $k \leq \ell$. Given a \kips\ $\s$, any $\x \in X$ with same first $k$ decisions as \s\ is said to be compatible with $\s$ (noted $\s \preccurlyeq \x$) and the subset of pipelines compatible with \s\ is noted $X(\s) = \{\x \in X; \s \preccurlyeq \x\}$.

A default distribution $\cal D$ is defined on $X$, involving a uniform distribution on all ${\cal A}_i$ and, conditionally to the selected $a_i$, a uniform distribution\footnote{Except for a few \hyper\ such as the number of selected features in feature selection, for which the default distribution is biased toward small values.} on the (bounded) $\Theta(a_i)$. The default distribution on $X(\s)$ is defined in the same way.

\subsection{Two intertwined optimization problems}
The difficulty lies in simultaneously tackling the structural optimization of ${\bf a}$ in ${\cal A}$ and the parametric optimization of the associated \hyper\ $\theta({\bf a})$ in $\Theta(\a)$ where i) the optimization objective is non-separable\footnote{That is, the marginal performance of $a_j$ depends on all other $a_k, k \neq j$ and on $\theta(\a)$. Likewise, the marginal performance of $\theta(a_j)$ depends on all $a_k$ and $\theta(a_k)$  for $k \neq j$.}; ii)
$\theta_j$ is of varying dimension, possibly depending on the value of some coordinates in $\theta_j$ (e.g. the number of neural layers controls the dimension of the neural layer size).
\def\hatF{{\mbox{$\widehat{\cal F}$}}}
At one extreme, one could optimize $\theta(\a)$ for every considered \a\ $-$ an obviously intractable strategy. At the other extreme, one could estimate the performance of \a\ from a few samples of $\theta(\a)$. 

\mosaic\ achieves an intermediate strategy: A surrogate model \hatF\ on $X$ is maintained, generalizing all computed performances; During the optimization of the pipeline structure with MCTS, when considering incomplete structural pipeline $\s \in {\cal A}_1 \times \ldots \times {\cal A}_k$, a full pipeline $\x$ such that $\s \preccurlyeq \x$ is determined along the line of Bayesian optimization  and the performance ${\cal F}(\x)$ is computed. Thanks to MCTS backpropagation step, this allows to build a proxy for the performance of $\s$.

More formally, the novelty in \mosaic\ is to tackle both structural and parametric optimization problems using two coupled strategies: MCTS is used to tackle the structural optimization of structure \a\ and Bayesian optimization is used to tackle the parametric optimization of $\theta({\bf a})$, where the coupling is ensured via the surrogate model(s). This hybrid strategy contrasts with that of \autosklearn\ (resp. most other AutoML approaches), optimizing both ${\bf a}$ and $\theta({\bf a})$ using Bayesian Optimization and a single surrogate model (resp. their own optimization methods). Note that in principle MCTS could be used to also achieve continuous optimization \cite{bubeck2011x}. However, the computational resource constraint on the AutoML problem, severely restricting the number of tree-walks, hinders a continuous MCTS optimization strategy.


\subsection{Partial surrogate models}
In \mosaic\ as in \as\ (section \ref{sec:soa}), a surrogate model \hatF\ of the optimization objective is built from all computed performances ${\cal F}(\x_u = (\a_u,\theta(\a_u)))$ . 

A first step is to derive from \hatF\ a surrogate model $\hatQ$ on pipeline structures. For $k < \ell$, let \s\ be a \kips, and let $\s.a$ denote the \kpips\ built from \s\ by selecting $a$ as $k+1$-th decision. Then the surrogate $\hatQ$ is defined as: 
\begin{equation}
\hatQ(\s, a) = \EE_{\x \sim {\cal D}[X(s.a)]} \left( \widehat{\cal F}(\x) \right) 
\approx \frac{1}{n_s} \sum_{j=1}^{n_s}  \widehat{\cal F}(\x_j)
\label{eq:hatQ}
\end{equation} 
estimated from a number $n_s$ ($n_s=100$ in the experiments) of configurations sampled in $X(\s.a)$.

A probabilistic selection policy $\pi$ can then be built from $\hatQ$, with: 
\begin{equation}
\pi(a|s) = \frac{\exp \left( \hatQ(s,a) \right)}{\sum_{b \in {\cal A}_k} \exp \left( \hatQ(s,b) \right) }
\label{eqnProba}
\end{equation}
Taking inspiration from \cite{AlphaGOZero}, this policy is used to enhance the MCTS selection rule (below).

\subsection{The \mosaic\ algorithm}\label{sec:playout}
\mosaic\ (Alg. \ref{euclid}) follows the general MCTS scheme (section \ref{mctsAlgorithm}), where the main four phases have been modified as follows: 
\paragraph{Down the MCTS tree} In a non-leaf node $\s$ of the MCTS tree, with $\s$ a \kips, the child node $a$ is selected in ${\cal A}_k$ using the \Zero\ criterion:\\
\begin{equation}
 \argmax_a \left( {\overline{Q}(\s, a) + \C * \pi(a|\s) * \frac{\sqrt{n(\s)}}{1+ n(\s.a)}} \right)
 \label{eq:mosaic}
\end{equation}

where $\overline{Q}$ is the median\footnote{The average was also considered, giving very similar results, except in rare cases of heavily failed runs.} of ${\cal F}(\x)$ for all $\x$ in $X(\s.a)$,  $\pi(a|\s)$ is defined by Eq.(\ref{eqnProba}), $n(\s)$ is the number of times $s$ was visited, and $\C$ is the usual constant controlling the exploration {\em vs} exploitation trade-off. 

\paragraph{Expansion} In a leaf node $\s$ of the MCTS tree, with $\s$ a \kips, the child node $a$ in ${\cal A}_k$ that maximizes the surrogate performance $\hatQ(\s,a)$ is added to the MCTS tree.  

\paragraph{Playout} Letting $\s$ be the (possibly complete) \kips, a full pipeline $\x$ with $\s \preccurlyeq \x$ is defined using a sampling playout strategy. Three sampling strategies were considered: i) a configuration is sampled according to the default distribution ${\cal D}(X(\s))$; ii) a local search 
around the best recorded pipeline $(\a^*,\theta^*)$ in $X(\s)$ is achieved and the best configuration according to $\widehat{\cal F}$ is retained; iii) a number of configurations is sampled after ${\cal D}(X(\s))$, together with a few configurations sampled via a local search around $(\a^*,\theta^*)$, and the sample \x\ that maximizes the Expected Improvement of  $\widehat{\cal F}$ is retained.
In all cases, the true performance ${\cal F}(\x)$ of the retained configuration is computed.

An empirical study (omitted for brevity) demonstrated that: the first sampling strategy is slow and prone to overfitting; the second strategy causes a loss of diversity of the considered pipelines, eventually resulting in a poor surrogate performance model $\widehat{\cal F}$. Hence only the third strategy is considered thereafter: the sampled configurations include $n_r$ ($n_r=1,000$ in the experiments) configurations sampled from default distribution ${\cal D}(X(\s))$, augmented with pipelines closest\footnote{Formally, one selects  every $(\a',\theta')$ such that either $\a'=\a^*$ and $\theta'$ differs from $\theta^*$ by a single \hypernots\ value; or $\a'$ differs from $\a^*$ by a single decision and $\theta'$ is the default \hypernots\ vector $\theta(\a')$.} to $(\a^*,\theta^*)$.   

\paragraph{Back-propagation} 
Performance ${\cal F}(\x)$ is back-propagated up the tree along the current path, updating the corresponding $\overline{Q}$ values.
Example $(\x,{\cal F}(\x))$ is added to the surrogate training set, and the surrogate performance model $\widehat{\cal F}$ is retrained anew. 

\paragraph{Stopping criterion} The algorithm stops after the computational budget is exhausted (one hour per dataset in the experiments).

\subsection{Initialization and Variants}\label{sec:variants}
The order of the decisions in the structural pipeline is key to the optimization: while MCTS yields asymptotic optimality guarantees, the discovery of good decisions can be delayed due to poorly informative or unlucky starts \cite{CoquelinM07}. Accordingly, the order of decisions in the structural pipeline is fixed, and the first decision made in the root node of the tree is the choice of the learning algorithm. Note that each learning algorithm has an associated default complete pipeline. 

\paragraph{\mosaic.Vanilla} The initialization proceeds as follows: For each learning algorithm ($\s = (a)$ with $a \in {\cal A}_1$), its default complete pipeline is launched, together with $\kappa$ (= 3 in the experiments) other pipelines sampled from $X(\s)$, and their associated performances are computed. The initial surrogate model $\widehat{\cal F}$ is trained from the set of all such $(\x,{\cal F}(\x))$ and $\hatQ(\emptyset, a)$ is initialized for $a$ in ${\cal A}_1$. 

\paragraph{\mosaic.MetaLearning} borrows \as\ its better informed initialization, where the first 25 configurations are the best recorded ones for each of the nearest neighbors of the current dataset, in the sense of the meta-feature distance (section \ref{metaAndEnsemble}). The next configurations are selected as in \mosaic.Vanilla, and the actual search starts thereafter. 

\paragraph{\mosaic.Ensemble} is similar to \mosaic.Vanilla, but returns the compound model defined as a weighted sum of the models computed along the AutoML search, using an online ensemble building strategy \cite{caruana2004ensemble}.


{\fontsize{10}{12}
\begin{algorithm}[H]
\caption{\mosaic\ Vanilla}\label{euclid}
\begin{algorithmic}[1]

\Procedure{Selection(state \s)}{}
\While{state not terminal}
    \State $a \gets$ Select action using Eq. 5
    \State \Return Selection(\s.a)
\EndWhile
\Return \s
\EndProcedure

\Procedure{Expansion(state \s)}{}
    \State \Return  $\argmax_{a}{\hatQ(\s,a)}$
\EndProcedure

\Procedure{Playout(state)}{}
    \State $P \gets {\cal D}[X(state)]\cup{Neighbor(x^*_l)}$ //  best configuration $x_i^* \in X(\textit{state})$
    \State \Return $\argmax_{c \in P}{EI(c)}$ // Expected improvement
\EndProcedure

\Procedure{\mosaic}{$T, d$}
    \While{$t < T$}
        \State $s \gets \emptyset$
        \State $s \gets$ Selection($s$)
        \State $a \gets$ Expansion($s$)
        \State x $\gets$ Playout($s \cup \{a\}$)
        \State Observe performance $r$ of $x$ on $d$
        \For{$p \in ancestors(s)$}
            \State Update $Q$ at state $p$ with $r$
            \State $n(p) \gets n(p) +1$
        \EndFor
    \EndWhile
\EndProcedure

\end{algorithmic}
\end{algorithm}
}



\section{Experimental Setting}\label{sec:goal}
\subsection{Goals of experiment} The empirical validation of \mosaic\ firstly aims to assess its performance compared to \as\ \cite{NIPS2015_5872}, that consistently dominated other systems in the international AutoML challenges \cite{guyon2015design}. The other AutoML system used as baseline is the evolutionary optimization-based\footnote{
\Alpha\ \cite{AlphaD3M} and \Autostacker\ \cite{AutoStaker} could not be considered due to lack of information.} \TPOT\ (v0.9.5) \cite{OlsonGECCO2016}.


The second goal of experiments is to better understand the specifics of the AutoML optimization problem. A first issue regards the exploration {\em vs} exploitation trade-off on the structural {\em vs} parametric subspaces and the merits of using MCTS as opposed to Bayesian optimization on the structural space. 
A second issue regards the  impact of the MetaLearning initialization. 
 MCTS is notorious to achieve a consistent though moderate exploration, which as said might slow down the search due to unlucky early choices. The smart initialization tends to prevent such hazards. On the other hand, if the initialization is {\em very} effective, the more conservative \as\ exploration strategy might be more appropriate. 

The exploration strategies of \mosaic\ and \as\ are compared, and the diversity of the visited configurations is examined in \cite{ArxivMosaic}.


\subsection{Experimental setting.}
\paragraph{Search space.} A fair comparison is ensured by assessing \as\ and \mosaic\ on the same \textit{scikit-learn} portfolio \cite{scikit-learn}. 
The search space involves 16 ML algorithms, 13 pre-processing methods, 2 categorical encoding strategies, 4 missing values imputation strategies, 6 rescaling strategies and 2 balancing strategies.\footnote{The reader is referred to \cite{ArxivMosaic} for more detail.} 
The size of the structural search subspace is 6,048 (due to parameter dependencies). The overall parametric search space has dimensionality 147 (93 categorical scalar \hyper, 32 integer, 47 continuous). Each \hypernots\ ranges in a bounded discrete or continuous domain. For each configuration $\x = (\a,\theta(\a))$, $\theta(\a)$ involves a dozen scalar \hyper\ on average. 

\mosaic\ involves 2 hyper-\hyper\ additionally to those of \as: the number $n_s=100$ of samples to compute $\hatQ$ (Eq. \ref{eq:hatQ}), $C_{ucb}=1.3$ controlling the exploration {\em vs} exploitation (Eq. (\ref{eq:mosaic})) and the coefficient of progressive widening $PW = 0.6$. Shared hyper-\hyper\ include: number $n_r$ of uniformly sampled configurations 
and variance $\epsilon=.2$ for the local search in the Playout phase (section \ref{sec:playout}). 


\begin{figure*}[t]
    \centering
    \includegraphics[width=\textwidth]{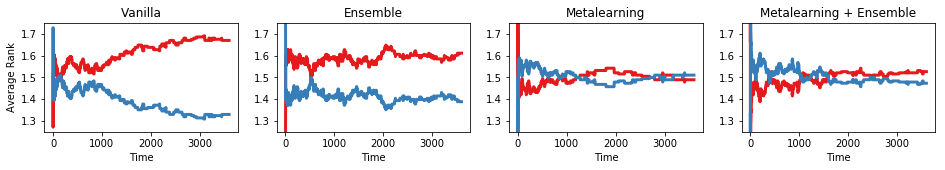}
    \caption{Comparative assessment of \textcolor{blue}{\mosaic} and \textcolor{red}{\as}: Average performance rank (the lower the better) on OpenML-100 {\em vs} CPU time of the Vanilla, Ensemble, MetaLearning and Ensemble+MetaLearning variants (left to right). Better seen in color.}
    \label{fig:comparison}
\end{figure*}

\paragraph{Benchmark suite} The compared AutoML systems are assessed on the OpenML repository \cite{OpenML2013}, including 100 binary and multi-class classification problems. 
The overall computational budget is set to 1 hour for each dataset. Computational times are measured on an AMD Athlon 64 X2, 5GB RAM.
For all systems, every considered 
$\x$ configuration  is launched to learn a model from 70\% of the training set with a cut-off time of 300 seconds, and performance ${\cal F}(\x)$ is set to the model accuracy on the remaining 30\%. After 1 hour, for each system the best configuration $\x^*$ is launched to learn a model on the whole training set and its performance on the (unseen) test set is reported. Finally, this performance is averaged over 10 independent runs, and the average is reported as the system performance on this dataset. For the Meta-Learning variant, the considered archive includes all datasets but the one under examination.

For each dataset, the performances achieved by all systems are ranked. The overall performance of a system is its average rank over all  (the lower the better).
As the rank indicator might be blurred when many systems and their variants are considered together, duels between pairs of systems (\mosaic.X against \as.X, where X ranges in {\em Vanilla, Meta-Learning, Ensemble, Meta-Learning+Ensemble}, section \ref{sec:variants}), are considered.


\section{Empirical Validation} \label{sec:expe}

\paragraph{Vanilla variants.} The comparative performances of Vanilla \as, \TPOT\ and \mosaic\ {\em vs} computational time are displayed on Figs. \ref{fig:comparison}-a and \ref{fig:vanilla}, showing that the hybrid optimization used in \mosaic\ clearly improves on the Bayesian optimisation-only used in \as\ (and on the evolutionary optimization-only used in \TPOT) from the early stages until the end. 

\definecolor{darkspringgreen}{rgb}{0.09, 0.45, 0.27}
\begin{figure}[b]
    \centering
    \includegraphics[width=.5\textwidth]{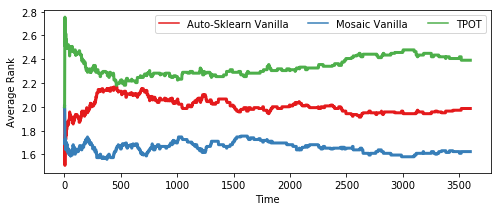}
    \caption{Average performance ranks (lower is better) on OpenML-100 {\em vs} CPU time of the Vanilla versions of \textcolor{blue}{\mosaic\ (bottom)}, \textcolor{red}{\as\ (middle)}, and \textcolor{darkspringgreen}{\TPOT\ (top)}. Better seen in color.} 
    \label{fig:vanilla}
\end{figure}

\begin{figure}[b!]
    \centering
    \includegraphics[width=.41\textwidth]{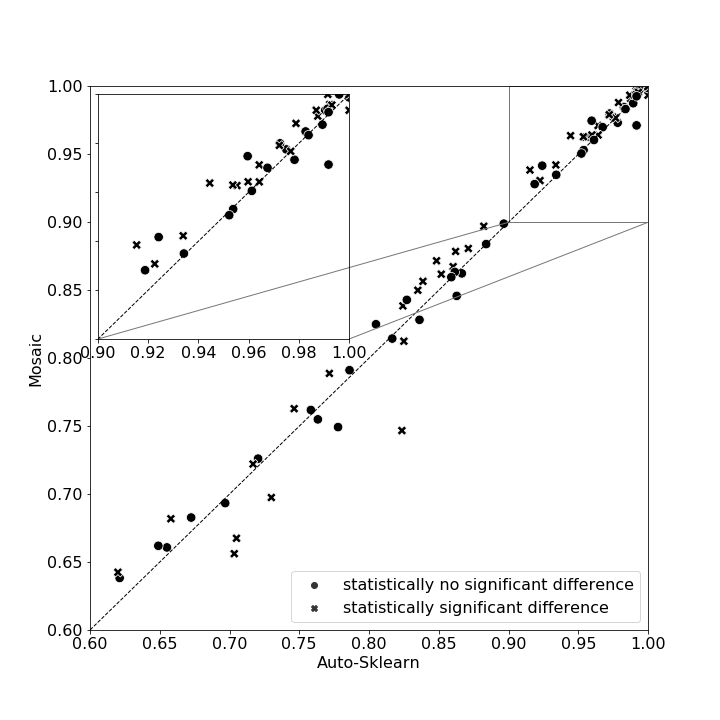}
    \caption{Performance of \mosaic\ (y-axis) versus \as\ (x-axis) on OpenML-100. Datasets for which the difference is statistically significant (resp. insignificant) after MWW test with confidence 5\% are represented with a $\times$ (resp $\bullet$).  } 
    \label{fig:diagonale}
\end{figure}
The actual performances of the configurations respectively selected by \as\ and \mosaic\ are reported on Fig. \ref{fig:diagonale}. According to a Mann-Whitney-Wilcoxon test with 95\% confidence, \mosaic\ significantly outperforms \as\ on 21 datasets out of 100; \as\ outperforms \mosaic\ on 6 datasets out of 100. 
Additionally, \mosaic\ improves on \as\ on 35 other datasets (though not in a statistically significant way), and the reverse is true on 18 datasets. Both are equal on 18 datasets and both systems crashed on 2 datasets.



\paragraph{MetaLearning and Ensemble variants.}
The impacts of the MetaLearning and Ensemble variants are displayed on Fig. \ref{fig:comparison}. While \mosaic\ dominates \as\ as long as the Vanilla variants are considered (Fig. \ref{fig:comparison}-a), the difference decreases for the Ensemble variant (Fig. \ref{fig:comparison}-b) and it becomes non-statistically significant for the MetaLearning variant (Fig. \ref{fig:comparison}-c), as well as for  the MetaLearning + Ensemble variant (Fig. \ref{fig:comparison}-d). 

A closer inspection of the results reveals that the best \as\ configuration is almost always found during the initialization and \as.MetaLearning thereafter mostly explores the close neighborhood of the initial configurations.
In the meanwhile, \mosaic\ follows a more thorough exploration strategy; this exploration might entail a bigger risk of overfitting, discovering configurations with better performance on the validation set, at the expense of the performance on the test set. 

\paragraph{Sensitivity w.r.t. \mosaic\ \hyper }
Complementary sensitivity studies have been conducted to assess the impact of \mosaic\ \hyper.
For computational reasons, only 30 datasets out of 100 have been considered, and 
\mosaic\.Vanilla is run 5 times with a 1 hour budget on each dataset.

\begin{figure}[ht]
    \centering
    \includegraphics[width=.4\textwidth]{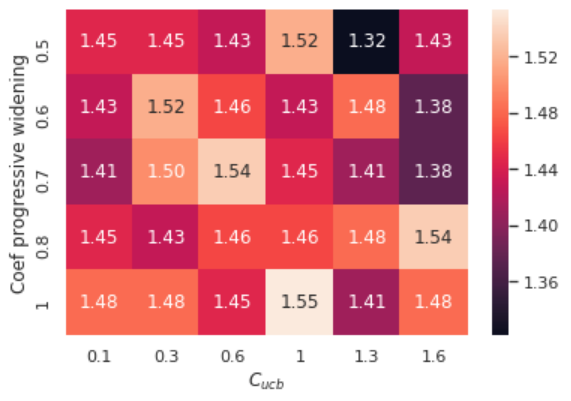}
    \caption{Sensitivity study w.r.t. \hyper\ $C_{ucb}$ and $PW$ (progressive widening in expansion phase), for $n_r = 100$: Average rank of \mosaic.Vanilla against \as.Vanilla (the lower, the better). Better seen in color (\mosaic\ in blue and \as\ in red).} 
    \label{fig:mcts_para}
\end{figure}

Fig. \ref{fig:mcts_para} displays the average rank of \mosaic.Vanilla at the end of the learning curve, for  $C_{ucb}$ in $\{.1, .3, .6, 1, 1.3, 1.6\}$ and PW in $\{1, .8, .7, .6, .5\}$, showing that \mosaic\ dominates Auto-Sklearn for 24 settings out of 30. 

\begin{figure*}[ht]
    \centering
    \hspace{-1.6cm}\includegraphics[scale=0.5]{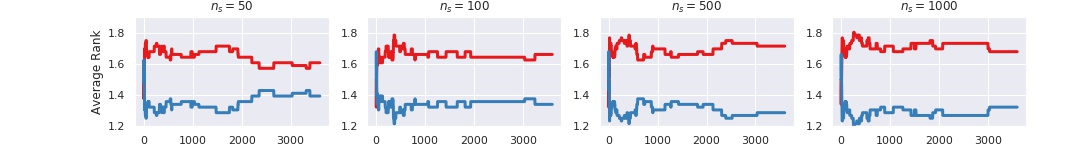}
    \caption{Sensitivity study w.r.t. \hyper\ $n_s$ for $C_{ucb} = 1.3$ and $PW = 0.6$: Average rank of \mosaic.Vanilla against \as.Vanilla. Better seen in color (\mosaic\ in blue and \as\ in red).
    } 
    \label{fig:mcts_nr}
\end{figure*}

Fig.~\ref{fig:mcts_nr} displays the average rank vs time of \mosaic.Vanilla for different values of $n_s$ (50, 100, 500, 1000), showing the low sensitivity of the performance w.r.t. $n_s$ in this range for $C_{ucb}=1.3$ and $PW=.6$.

\paragraph{Comparing \mosaic\ and \as\ exploration of the search space}\label{sec:explo}

The differences in the exploration strategies of \as\ and \mosaic\ become more visible at a later stage of the search:  \mosaic\ switches to the exploitation of the most promising MCTS subtrees (subspaces of the search space) and avoids regions where the last visited configurations were bad; on the other hand, \as\ continues to explore even if the sub-space includes quite a few bad configurations \cite{ArxivMosaic}.

\section{Discussion and Perspectives}
The main contribution of the paper is the new \mosaic\ scheme, tackling the AutoML optimization problem through handling both the structural and the parametric optimization problems. The proposed approach is based on a novel coupling between Bayesian Optimization and MCTS strategies, that are tied by sharing the same surrogate model. In MCTS the surrogate model is used to estimate, in all nodes, the average performance of all subtrees (ends of pipeline) below this node, and thus to choose the next node. The same surrogate model is used during the roll-outs, to choose the optimal hyper-parameters of the pipeline using a Bayesian Optimization strategy.

Empirically, the results demonstrate that \mosaic\ significantly outperforms the challenge winner \as\ on the OpenML benchmark suite, at least as long as the Vanilla and Ensemble variants are considered. With the MetaLearning variant however, the difference becomes insignificant as the bulk of optimization is devoted to the initialization (all the more so for large datasets, due to the one hour cut-off time).  

The limitation of such a smart initialization is twofold. On the one hand, it relies on preliminary expensive computations to build the archive (one day computation {\em per} dataset on OpenML-100); on the other hand, it assumes the representativity of the problems in the archive.
On-going work is concerned with estimating the risk of overfitting the OpenML benchmark, through measuring the sensitivity of the \as\ and \mosaic\ MetaLearning variants when varying the fraction of the datasets in the archive. 

In any case, the experimental evidence suggests that Vanilla \mosaic\ offers a robust and efficient AutoML facility when tackling a new application domain, and/or in the absence of a comprehensive archive. 

A long term research perspective is to reconsider the design of the meta-features \cite{KateSmithMiles}. In principle, a binary classification problem can be associated to any ML algorithm, where a dataset belongs to class $+$ if the algorithm performs comparatively well on this dataset, and class $-$ otherwise. The perspective is to apply equivariant learning \cite{equivariant2016} at the dataset level to tackle this binary classification problem, and use the resulting equivariant classifier as (cheap) meta-feature. 


\section*{Acknowledgments}
This work was funded by \href{https://www.ademe.fr/next}{the ADEME \#1782C0034 project {\em NEXT}}. 
\bibliographystyle{named}
\bibliography{ijcai19}
\end{document}